\documentclass[11pt]{article}
\usepackage{coling2020}
\usepackage{times}
\usepackage{url}
\usepackage{latexsym}
\usepackage{blindtext}
\usepackage{algpseudocode}
\usepackage{mathptmx}
\usepackage[linesnumbered,ruled,vlined]{algorithm2e}

\usepackage{graphicx}
\usepackage{wrapfig}
\usepackage{enumitem}
\usepackage{fancyhdr}
\usepackage{amsmath}
\usepackage{index}
\usepackage{tikz}
\usepackage{caption}
\usepackage{subcaption}

\colingfinalcopy 

\title{[Modelling Causal Reasoning in Language: Detecting Counterfactuals] at SemEval-2020 Task [5]: [Counterfactual Detection meets Transfer Learning ]}

\author{Nwaike Kelechi \\
  Xidian University\\
School of Artificial Intelligence\\
  {\tt kell21n@gmail.com} \\\And
  Licheng Jiao\\
  Xidian University\\ 
School of Artificial Intelligence \\
  {\tt lcjiao@mail.xidian.edu.cn}\\}

\date{}

\begin{document}
\maketitle
\begin{abstract}

    We can consider Counterfactuals as belonging in the domain of Discourse structure and semantics\cite{pra}, A core area in Natural Language Understanding and in this paper, we introduce an approach to resolving counterfactual detection as well as the indexing of the antecedents and consequents of Counterfactual statements. While Transfer learning is already being applied to several NLP tasks\cite{raf}, It has the characteristics to excel in a novel number of tasks. We show that detecting Counterfactuals is a straightforward Binary Classification Task that can be implemented with minimal adaptation on already existing model Architectures, thanks to a well annotated training data set,and we introduce a new end to end pipeline to process antecedents and consequents as an entity recognition task, thus adapting them into Token Classification.

\end{abstract}

\section{Introduction}
\label{intro}
\par While Counterfactuals have been studied in different ways and in various domains, such as in Interpretable Machine learning\cite{mol}, Philosophy\cite{Col},Mathematics\cite{Cor}, The research in this paper is concerned with the Detection of Counterfactuals as they occur in the domain of Natural Language and specifically, in English Language. We can express a description of Counterfactuals in this manner: They have a basic building block which we shall refer to as the antecedent,which denotes an event that did not occur, but which might have brought about another event, which we refer to as the consequent, if it did occur. More simply, \emph{If antecedent, then consequent where the antecedent is usually false}\cite{Gin}.The dual task of detecting counterfactuals and locating the start and end indexes of the antecedent and consequent leverages on the data set and boundaries as defined by\cite{yan} for the SemEval task 5.
\par  Transfer learning is the transference of learner knowledge from one domain to the other, and if it is to be relevant, the domains need to be related\cite{Weis}. Intuitively, we do not fine-tune a model trained on programming language corpus for prediction with a movie reviews data set.Transfer learning becomes even more useful when there is inadequate data for training.\cite{pan}, and this is a situation that occurs all too often as preparing large data sets is an intensive task.\cite{surya} compared classification model accuracy with a range of data sizes, and while the effect of data set size on model accuracy is dependent on the model in use as well as the specifics of the classification task, it was shown that using the BERT language model with the IMDB reviews data set, there was only a 5.3 percent degradation in accuracy when trained with 500 versus 22500 samples.There is however, such a thing as too little data, as a train size of 125 samples suffered a 24 percentage point drop in accuracy versus 500 samples.

 \par Counterfactuals occur in only about 1\% of tweets\cite{son}, and that highlights their infrequency and thus the focus of our system in this two-track task is to introduce a pipeline that leverages transfer learning on extant pre-trained language models specifically to resolve the tasks of Counterfactual detection, which due to their characteristic of possessing implicit and explicit forms\cite{and}, limits the capability of rule-based or statistical methods to accurately detect some forms of counterfactuals. 

  With the conviction that the detection of Counterfactuals is a Sentence level classification task- by virtue of the provided data sets, and the identification of the antecedents and consequents a word level classification task, We introduce in this paper two pipelines for both tasks.The underlying language models used are base models of BERT\cite{dev}and RoBERTa\cite{liu}, so that following the trend of the large models outperforming the base models\cite{dev} in sentence and token classification tasks, there is room for more qualitative optimization. All of the code used for the tasks as well as a basic guide are available on github. \footnote[1]{ https://github.com/Kc2fresh/Extracting-Counterfactual-data/}

\section{Background}

\begin{enumerate}

\item A snippet of the data sets is shown in Table \ref{table:data1} and \ref{table:2}, the data sets are statements in the English language, possibly of the news genre. 1 represents Counterfactual, and 0 for Non-Counterfactual. A detailed description of the train and test sets for the two tracks  is shown in Table \ref {table:noe} and \ref{table:nop}.  There has been a previous attempt at detecting  counterfactuals, albeit in social media texts, by\cite{son}, using a combination of rules and a supervised classifier, with a data set built with a combination of rule-based keywords and manual annotation. In this task, we directly skip any attempt at rules or probabilistic methods,  even if it may have been tempting to get fixated on \emph{modal verbs} and words like \emph{if}and \emph{wish} see Table\ref{table:data1}, instead for sub-task 1, fine-tuning the the language models, already pre-trained on large amounts of unsupervised text, with the train set.

\begin{table}
\begin{tabular}{l|l}
\hline
 1 & If only they had adhered to conservative principles...things would have been DIFFERENT.\\
 0 & If needed, I would like to have the right to try.\\
\hline
 \end{tabular}
 \caption{Task 1:Data Sample}
 \label{table:data1}
\end{table}

\begin{table}
\begin{tabular}{l|l|l|l|l}
\hline
Sentence& AS-ID & AE-ID & CS-ID & CE-ID\\[0.2ex]
\hline
I just wish it had been my hand holding my daughter, not his.& 0 & 50 & -1 & -1\\
\hline
\end{tabular}
\caption{Task 2:Data Sample: A=antecedent, C=Consequent, S-ID=start index, E-ID=end index }
\label{table:2}
\end{table}

\begin{table}[ht]
   \begin{subtable}[h]{0.45\textwidth}
   \small
     \centering
     \begin{tabular}{l|l|l|l}
     \hline\hline
      Task-1  & Train set   & Percentages  &  Test set\\[0.5ex]
      \hline
      CF      &  1454       &  12.6\%       &         \\      
      \hline
      Non-CF  & 11546       &  87.4\%       &         \\
      \hline
      Total   & 13000       & 100\%        &   7000  \\
      \hline
      \end{tabular}
      \caption{ Task 1: Train\&Test sets}
      \label{table:noe}
      \end{subtable}
       \hfill
     \begin{subtable}[h]{0.45\textwidth}
     \centering
     \begin{tabular}{l|l|l}
     \hline\hline
     Task-2  & Train set  &  Test set\\[0.5ex]
     \hline
      CF      &  3551      &  1950\\      
     \hline
     \end{tabular}
     \caption{Task2 : Train\&Test sets}
     \label{table:nop}
     \end{subtable}
     \caption{Train\&Test sets}
     \label{table:nonl}
\end{table}

\end{enumerate}




\section{System Overview}
 \par We investigated two forms of Transfer learning, feature extraction and fine-tuning\cite{pet}, where fine-tuning was found to give slightly better results than feature extraction, particularly on the transformer models on which our experiments on this task is based.
  The second modelling decision of the system was dictated by computational restrictions, and so for both sub-tasks, the choice was made to use the base language models. However, the inference code can with minor tweaks, be adapted to newer and larger models for improved performance. Both underlying language models fine-tuned for this task, are multi-layer, bidirectional Transformer encoders based on the transformer Architecture\cite{vas}, and trained on large amounts of unsupervised data, thus bringing the possibility of effective supervised transfer learning to NLP\cite{raf}. Specifically with the BERT and RoBERTa models, for fine-tuning tasks, a classification layer is added above the pre-trained model, with classification layer weights introduced in addition.
\par  The first sub-task is defined as a binary label sequence classification problem, and the data set is tokenized in sentence batches as a sequence task, while the second sub-task is defined as a word level inference task, and thus as a token classification task.
\subsection{Sub-task1}
  \par The experiments were done in Python, using Google Colab, with the Hugging-face implementation of the models in Pytorch. The default BERT model tokenizers were used in each instance, and maximum length was set to 128. Three instances in the data set exceeded a maximum sequence length 512, and were truncated without trouble as they represent a small fraction of the total data set. We experimented with batch sizes of 8, 16 and 32, and fell back to 32, as it gave the best accuracy on the validation set. 

\subsection{Sub-task 2}
\par Framing the task as a token classification problem, on par with an entity recognition problem, required an arrangement of the input data similar to an NER data set. Two different pre-trained models were used for the sub-tasks. In all instances, the span of the antecedents and consequents were tested against the full sentences and all the elements in the span were replaced by the tags \emph{ante} and {\emph{cons} or \emph{0}, depending on which situation applied. The sentences in each row were stacked with one word per row along with the matching tag. The Part of speech tags were acquired with spaCy \footnote[1]{ https://www.spacy.io}but due to tokenizer incompatibilities, were discarded and NLTK\footnote[2]{ https://www.nltk.org} was used instead as it was able to apply POS tags on a word by word basis. However, this method is known to be less accurate.  For fine-tuning, we generally followed the hyper-parameters as recommended in the model papers. Most model hyper-parameters are the same as in pre-training, with the exception of the batch size, learning rate, and number of training epochs. The dropout probability was always kept at 0.1. The models for both sub-tasks were downloaded from Hugging-face.\footnote[3]{ https://github.com/huggingface}

\DontPrintSemicolon
\newcommand{\To}{\mbox{\upshape\bfseries }}


\begin{algorithm}
  \caption{Pipeline for Sub-task-2}
  \textbf{Convert Data set to NER style format for Token Classification }\;
  \ Replace each antecedent and consequent span of the sentence with corresponding label\;
  \ Optionally, Use NLTK to build POS tags\;
  \ Stack the data set on a Word level as shown in Table \ref{table:total} \;
  \textbf{Train the Tokens and perform Inference}\;
  \textbf{Post-processing and locating the indexes}
\end{algorithm}

\begin{table}[ht]
   \begin{subtable}[h]{0.45\textwidth}
      \centering
       \begin{tabular}{l l l}
       \hline
        Word  &  Tag&\\[0.5ex]
        \hline
        If &  ante&\\      
        \hline
         I & ante&\\
         \hline
         had & ante&\\
         \hline
         10  & ante\\
         \hline
         pharmacists & ante\\
         \hline
         . & .\\
         \hline
         . & .\\
         \hline
         people & cons\\
         \hline
         more & cons\\
         \hline
         effectively  & cons\\
         \hline
         \end{tabular}
         \caption{ Task-2 Dataset}
         \label{table:nonlie}
      \end{subtable}
       \hfill
     \begin{subtable}[h]{0.45\textwidth}
        \centering
         \begin{tabular}{r r r}
         \hline
          Word  &  Tag & POStag\\[0.5ex]
          \hline
           If &  ante & IN\\      
           \hline
           I & ante & PRP\\
           \hline
           had & ante&VBD\\
           \hline
           10  & ante&CD\\
           \hline
           pharmacists & ante&NNS\\
           \hline
            . & .&\\
            \hline
            . & .&\\
            \hline
            people & cons&NNS\\
            \hline
            more & cons&RBR\\
            \hline
            effectively  & cons&RB\\
            \hline
            \end{tabular}
            \caption{ Task-2 Dataset with POS tags}
            \label{table:nonlingo}
        \end{subtable}
    \caption{Training pre-process for Sub-task 2}
    \label{table:total}
\end{table}

\section {Experimental Setup}

\subsection{Data Pre-processing}
\begin{enumerate}
 \item Different approaches for Text Pre-processing for NLP tasks exist\cite{dud}, depending on the task in question. Our system design choice was to use no pre-processing for the basic model, given as the BERT and RoBERTa models were trained on unsupervised text. For ablation purposes, an alternate data set was stripped of punctuation, rare characters and hashtags and also trained. In the same vein, Data augmentation using Back-translation with Chinese as the intermediary language, and 1000 statements which glaringly retained their counterfactual labels were grabbed. There has been some research on the usefulness of Data Augmentation to NLP tasks.\cite{wei} proposed techniques of data Augmentation for NLP tasks and \cite{Kob}had previously demonstrated the benefits of data augmentation, in situations of limited data sets or small training samples. With a train test size of 13,000 however, it was considered that only large scale data augmentation would bring appreciable improvements\cite{surya}. Nevertheless,we included another 1000 counterfactual back-translated sentences to an alternate data set. It was not expected to have much effect and there was the real risk with data augmentation of muddying the labels and so the chosen augmented statements were looked over briefly to ascertain that the data set was not being degraded.
\item Sub-task 2 was interpreted as a word level classification task, and thus similar to an entity recognition task. No data pre-processing was done for this task, except the preparation of the data set for the model pipeline. The data preparation required the data set to be re-arranged on a word level along with its matching tag, similar to a NER data set. The script to do this is available along with the rest of the code. 
 
\end{enumerate}
\subsection{Model Fine-Tuning}
\begin{enumerate}
\item Fine-Tuning of the model for sub-task 1, was tried multiple times and with various test-splits and hyper-parameters to get a feel of the optimal settings. All the base un-cased, pre-trained BERT and RoBERTa Sequence classification models used for predictions were trained with a 95/5 split over 3 epochs,batch size = 32, optimized with AdamW, with an initial learning rate of 5e-5, dropout = 0.1 and eps = 1e-8. All experiments were performed in Google Colab and were based on Hugging face and Pytorch Implementations of the Transformer models.  
\item The model Fine-Tuning for sub-task 2,was the BERT-base cased, with a token classification linear layer on top of the hidden-states output, The cased model was used specifically because this is an entity recognition task, the test splits was 90/10, batch size = 32, optimized with AdamW, learning rate of 3e-5, and trained over 4 epochs. The experiment was implemented in Pytorch using the GPU made available in Google Colab.   
\end{enumerate}

\subsection{Post-processing}
\begin{enumerate}
\item No significant post processing tasks were required for sub-task 1, the labels were gathered into a list and put in a data frame to match with the sentence ID's for onward submission.
\item For sub-task 2, the tokenization mis-match as a consequence of our choice of underlying language model presented unique problems when trying to retrieve the span as the tokenized output labels needed to be detokenized back in such a way that the predicted labels matched the initial input test sentences so as to extract the indexes. For example, given a word \emph{if} as the output word where the antecedent span begins, There were situations were multiple \emph{if's} were in a sentence, so that a basic attempt at string replace would introduce unexpected inaccuracy. While there can be better approaches to this, and work will continue to improve it, the code for the best run is included with the publicly available code.

\begin{table}[ht]
\small
\begin{center}
\begin{tabular}{lllllllllllllllllll}
\hline\hline
[CLS]&If&,&during&2012&,&you&had&invested&in&the\\[0.1ex]
\hline
'0'&'ante'&'ante'&'ante'&'ante'&'ante'&'ante'&'ante'&'ante'&'ante'&'ante'\\  
\hline
S&\&&P&500&,&your&investment&would&have&returned\\
\hline
ante'&'0'&'ante'&'ante'&'0'&'cons'&'0'&'cons'&'cons'&'cons''\\  
\hline
15&.&9&\%&,&after&factoring&in&dividends&.&[SEP]\\
\hline
'cons'&cons'&'cons'&'0'&'0'&'0'&'0'&'0'&'0'&'0'&'0'&\\
\hline
\end{tabular}
\caption{ Task-2: Output 'ante' and 'cons' tags for a Sample statement}
\label{table:nol}
\end{center}
\end{table}

\end{enumerate}


\section{Results}
\begin{enumerate}
\item The highest F1 score achieved in the experiments was a RoBERTa base model see Table \ref{table:results}, with a score putting it 0.04 behind the leading F1 of 0.909 on both the official and unofficial leaderboard, and 0.01 ahead of the highest score achieved with BERT base.The result of a tail Ensemble voting\cite{Xu} returned a slightly worse score than the best Roberta score, but this is within the margin of error. 
  Training with a data set cleaned of punctuation did not show improved benefits,although a closer analysis would show if any, the results were slightly worse. The data augmented model with 1000 back-translated Counterfactual statements inspected for did not yield any benefits either, but it is not clear whether this can be attributed to the minimal augmentation performed.
  A look at the recall and precision scores, indicates a variance along both sides, and not sufficient for generalization, but are indicative of the strength of the underlying models and the suitability of counterfactual detection as a transfer learning task.

\begin{table}[ht]
\centering{}
\
\begin{tabular}{l|l|l|l|l}
\hline\hline
Model & Method  & F1 & Recall& Precision \\[0.5ex]
\hline
Bert-base(uncased)&Best of 3 runs &0.859 & 0.836 & 0.884\\

& pre-processed train\&test set & 0.841 &0.897 & 0.791\\

& + BT data &  0.854 &0.841 & 0.866\\     
\hline
RoBERTa base(uncased)& Best of 2 runs &  0.869 &0.881 & 0.858\\     

& pre-processed train\&test set &  0.860 &0.846 & 0.874\\    
\hline
Ensemble Voting& RoBERTa + BERT &  0.868 &0.890 & 0.848\\  
\hline 
\end{tabular}
\caption{Task-1 Results}
\label{table:results}
\end{table}

\item The three official results of sub-task 2 are interesting in that while neither are high ranking on the official metrics,are all post-processed outputs of the same predicted model, As seen from the output of the model in Table \ref{table:nonl}, the predictions are generally relatively impressive,although there was sometimes a tendency to not recognize a pronoun as the beginning a span, but with the wide variance in scores given that they are outputs of the same model, show that the post processing pipeline is inefficient and can be improved.

\begin{table}[ht]
\centering{}
\
\begin{tabular}{|l|l|l|l|l|}
\hline\hline
Type & F1 & Recall& Precision & Exact Match\\[0.5ex]
\hline
Step A & 0.648 & 0.690 & 0.673 & 0.000513\\
\hline
Step B & 0.551 & 0.514 &0.567 & 0.000513\\
\hline
Step C & 0.667&  0.728 &0.666 & 0.0\\     
\hline
\end{tabular}
\caption{\label{table:nonliner} Task-2 Results}
\end{table}

\end{enumerate}

\section{Conclusion}
In this work, we applied transfer learning using language models to Counterfactual detection, as well as the indexing of the antecedents and consequents, achieving an F1 score in sub-task 1, only slightly less than the top score, even while using base language models. It is demonstrated that the primary factor in the improved detection over previous probabilistic and rule-based methods is the underlying pre-trained language models. Thus higher accuracy will be achieved with improvements to the underlying models and In task 2, further development needs to be done on the sub-task pipeline, to handle output more efficiently, and achieve results that do not diminish the output of the language model prediction pipeline.


{}

\end{document}